\documentclass[letterpaper]{article} % DO NOT CHANGE THIS
\usepackage{aaai25}  % DO NOT CHANGE THIS
\usepackage{times}  % DO NOT CHANGE THIS
\usepackage{helvet}  % DO NOT CHANGE THIS
\usepackage{courier}  % DO NOT CHANGE THIS
\usepackage[hyphens]{url}  % DO NOT CHANGE THIS
\usepackage{graphicx} % DO NOT CHANGE THIS
\usepackage{natbib}  % DO NOT CHANGE THIS AND DO NOT ADD ANY OPTIONS TO IT
\usepackage{caption} % DO NOT CHANGE THIS AND DO NOT ADD ANY OPTIONS TO IT
\usepackage{algorithm}
\usepackage{algorithmic}
\usepackage{amsmath}
\usepackage{amsthm}
\usepackage{amsfonts}
\usepackage{subcaption}
\usepackage{newfloat}
\usepackage{listings}
\DeclareCaptionStyle{ruled}{labelfont=normalfont,labelsep=colon,strut=off} % DO NOT CHANGE THIS
\floatstyle{ruled}
\newfloat{listing}{tb}{lst}{}
\floatname{listing}{Listing}
\nocopyright
\title{Does GPT Really Get It? A Hierarchical Scale to Quantify Human and AI's Understanding of Algorithms}
\author{Mirabel Reid\textsuperscript{\rm 1} and Santosh S. Vempala\textsuperscript{\rm 1}}
\affiliations {
    \textsuperscript{\rm 1} Georgia Institute of Technology, School of Computer Science\\
    mreid48@gatech.edu, vempala@cc.gatech.edu
}

\newtheorem*{definition}{Definition}
\begin{document}

\maketitle

\begin{abstract}
As Large Language Models (LLMs) are used for increasingly complex cognitive tasks, a natural question is whether AI really {\em understands}. The study of understanding in LLMs is in its infancy, and the community has yet to incorporate research and insights from philosophy, psychology, and education. Here we focus on understanding {\em algorithms}, and propose a hierarchy of levels of understanding. We validate the hierarchy using a study with human subjects (undergraduate and graduate students). Following this, we apply the hierarchy to large language models (generations of GPT), revealing interesting similarities and differences with humans. We expect that our rigorous criteria for algorithm understanding will help monitor and quantify AI's progress in such cognitive domains.   
\end{abstract}

\section{Introduction}

Since the release of GPT-4, mainstream users have begun to experiment with Large Language Models (LLMs) on increasingly complex tasks. However, the degree to which it is safe, legal, and ethical to rely on LLMs has been under fierce debate. Across many studies, researchers have identified apparent shortcomings of LLMs including hallucinations, inability to plan, and lack of understanding~\cite{rawte2023survey, mahowald2024dissociating, valmeekam2023planning}. However, the literature notably lacks rigorous criteria to measure the progress toward solving these issues.  A particular problem lies in claims surrounding understanding; AI understanding is frequently compared to human understanding, and it is folklore among AI researchers that the reasoning processes of LLMs differ from those of humans. While the concept of understanding is widely discussed, it remains ill-defined.

In this paper, we propose an {\em precise definition of understanding an algorithm} with the following properties: (a) it provides a scale by which to evaluate any entity's understanding of an algorithm, (b) it aligns with the standard usage of the term ‘understanding’ in philosophy and psychology, and (c) it can be used to evaluate AI's progress toward understanding algorithms.

\subsection{Motivation: Why Study Algorithm Understanding?}
Large language models are increasingly trusted for coding assistance. Code generation tools such as GitHub Copilot~\cite{GitHub} and Meta's Code Llama~\cite{roziere2023code}  are currently used in practice to improve developer productivity~\cite{vaithilingam2022expectation,mozannar2024reading} and assist novice programmers in learning~\cite{kazemitabaar2023studying,becker2023programming}. It is likely that the degree of AI involvement in software development will only grow as these tools improve. However, reliance on imperfect systems comes with risk. Tools such as Copilot are known to generate code that is subject to license~\cite{becker2023programming} or contains security vulnerabilities~\cite{pearce2022asleep}. The question of whether LLMs demonstrate meaningful understanding of algorithms is relevant if we are relying on LLMs to implement algorithms in production or teach them to novice programmers.

Algorithm understanding is distinct from language understanding and  deserves its own line of study. Those who argue that LLMs do not understand language draw a distinction between linguistic form and meaning\cite{bender2020climbing, mitchell2023debate, pavlick2023symbols}. When humans understand language, their understanding is informed by their communicative intent and the real-life properties of the objects described. Thus, a system trained only to replicate statistical correlations between words cannot understand language in the way that humans do. Algorithms, however, can be precisely represented using formal programming languages. One might argue that a computer can meaningfully observe an algorithm in full through code implementations and examples.

%That is not to say that a human's understanding of algorithms is not informed by real-world experience. For example, a student may understand shortest-path algorithms partially through their experience using navigational apps to get to work. Nonetheless, it is easier to work around this discrepancy between humans and LLMs when the object of study is formally defined. 
\subsection{Related Work}
\paragraph{Cognitive Abilities of LLMs.} The past few years have seen an explosion of studies exploring the ability of LLMs to answer complex mathematical questions. Researchers have developed prompting strategies to enable multi-step reasoning~\cite{wei2022chain, fu2022complexity}. Still others fine-tune models to improve mathematical problem-solving~\cite{yu2023metamath, luo2023wizardmath}. The benchmarks for these methods typically include large datasets such as GSM8k~\cite{cobbe2021gsm8k} (grade school word problems) and MATH~\cite{hendrycks2021measuring} (math competition problems). These works focus on correct evaluation and do not address whether the language models understand mathematical reasoning. 

Others have studied metacognitive skills in LLMs. \citet{didolkar2024metacognitive} investigate whether LLMs can assign skill labels to mathematical problems. Also related is \citet{aher2023using} which proposes Turing experiments comparing humans and LLM simulations. 

\paragraph{Understanding in LLMs.} A parallel line of work investigates language understanding in LLMs. A key concept in the debate over language understanding is the difference between linguistic \textit{form} and \textit{meaning}~\cite{bender2020climbing, merrill2021provable}. \citet{bender2020climbing} argue that an AI trained only on linguistic form (i.e. text) cannot understand meaning. In an opinion piece, \citet{pavlick2023symbols} counters this perspective, arguing that it is premature to draw conclusions on whether LLMs can model language understanding when the study of language models is itself in its infancy. There has been some effort to determine the extent to which LLMs represent linguistic meaning, primarily by studying word representations \cite{li2021implicit, patel2021mapping}. For a survey on linguistic competence in LLMs, see \cite{mahowald2024dissociating}. Also see \cite{mitchell2023debate} for a general survey on the debate over understanding. 

\paragraph{Theories of Understanding.} The debate over what constitutes understanding has a long history in philosophy and psychology. It is generally agreed that understanding is different from `mere' knowledge, but the nature of that distinction is up for debate~\cite{pritchard2009knowledge, baumberger2016understanding, paez2019pragmatic}. \citet{pritchard2014knowledge} provides some examples of when the concepts of `knowing why' and `understanding why' may not overlap. \citet{khalifa2017understanding} and \citet{baumberger2016understanding} are accessible surveys of this debate.

The philosophy of science also relates understanding and explanation, and the goal of explanation can be thought of as the production of understanding~\cite{friedman1974explanation, grimm2010goal, baumberger2016understanding}. \citet{wilkenfeld2016depth} argue the converse; they relate understanding to explanatory depth and claim that we attribute understanding in order to identify experts to consult. \citet{woodward2005making} overviews what defines a causal explanation. 

Also relevant to this work is the distinction between \textit{deep} and \textit{surface-level} learning from educational psychology~\cite{marton1976qualitative,beattie1997deep}. Perhaps the most influential categorization of educational goals is Bloom's Taxonomy~\cite{bloomtaxonomy}. This taxonomy has been revisited many times since its publication; notably, \citet{mayer2002rote} categorized student learning into cognitive processes and identified testable skills which arise with understanding. 
%This categorization partially informed the levels we define below. 
\section{A Definition of Understanding}
	 	 	 	
We ask the question: {\em how well does an entity understand an algorithm?} Our goal is a definition of understanding that is itself  algorithmically testable. Therefore, we adopt a functional lens, meaning that we define understanding by what it allows the entity to do. 

\subsection{Preliminaries}
In this work, we ask whether an entity $\mathcal{E}$ understands a computable function $f: \Omega\rightarrow \Sigma^*$. By {\em computable}, we say that there exists a Turing Machine which takes $x\in \Omega$ as input and halts with $f(x)$ on its tape, using a standard definition~\cite{sipser1996introduction}. Let $\mathcal{A}$ be an algorithm that computes $f$.  

We assume that an entity $\mathcal{E}$ (a) has a long-term memory system and (b) can perform computation, enabled by a working memory with finite capacity $M_\mathcal{E}$. We say that $\mathcal{E}$ \textit{knows} a function $f$ or algorithm $\mathcal{A}$ if it has a representation $R_f/R_{\mathcal{A}}$ stored in its long-term memory. When $\mathcal{E}$ computes the function $f$ on an input $x\in\Omega$, it runs an internal algorithm $\mathcal{A}_{\mathcal{E}}$, which may or may not be the same as $\mathcal{A}$. The entity's understanding of $\mathcal{A}$ will be measured by its ability to manipulate this representation to produce answers to queries. This definition is based on the Understanding as Representation Manipulability system (URM) proposed by Wilkenfield~\cite{wilkenfeld2013understanding}. 

For a particular input $x$, let $M(x)$/$T(x)$ be the memory/time required to evaluate the algorithm on $x$. 

The \textit{execution path} of $\mathcal{A}$ on an input $x$ is the sequence of states taken by the algorithm when executing on $x$. The \textit{trace} of the execution is the execution path plus the contents of the tape at each step. Finally, define a \textit{property} to be a function mapping the trace or execution path to $\{0,1\}$.
\subsection{Internal Representations} 
For our hierarchy, we employ the framework of Understanding as Representation Manipulability (URM)~\cite{wilkenfeld2013understanding}. This theory posits that understanding arises from the ability to modify the internal representation of a concept in order to make effective inferences. 

 For language models, the representation of a concept (such as an algorithm) is collected from the thousands of examples, explanations, and code snippets that appear in its training data. The mechanism behind human memory is not understood as precisely. However, humans also learn via hearing explanations, collecting examples, and reinforcing their knowledge. This forms a representation encoded in the neural pathways of our brains~\cite{durstewitz2000neurocomputational, gyorgy2019brain}. The goal of our hierarchy will be to test how the existing internal representation can be manipulated to produce responses at different levels of difficulty.  

In his description of URM, Wilkenfeld declines to characterize the structure of the representations, other than to state that they are ``computational structures with content that are susceptible to mental transformations''\cite{wilkenfeld2013understanding}. Thus, the definition of understanding is independent of the entity's internal structure and the mechanism for inference. This is in line with our functional lens; the level of understanding is based on the entity's ability to manipulate its representation to perform tasks at different levels of difficulty.

\subsection{Levels of Understanding}
\label{sec:levels}
\begin{figure}[h]
 \centering
    \includegraphics[width = 0.8\columnwidth]{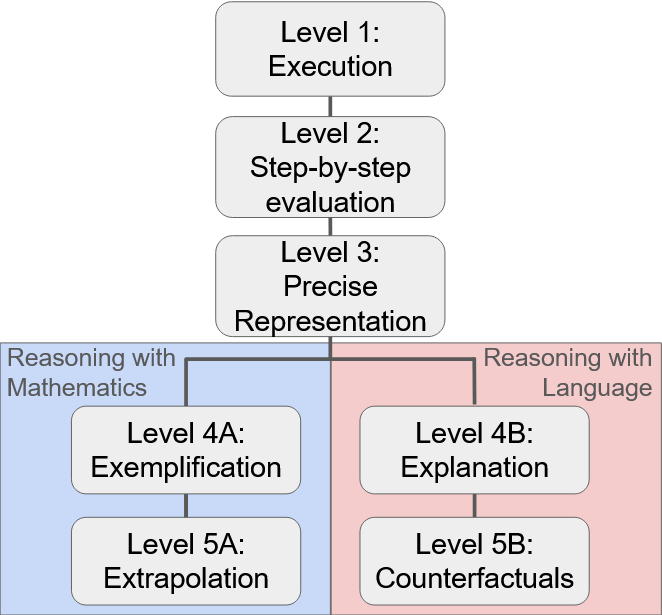}
	\caption{A hierarchy of understanding.}
	\label{fig:levels}
\end{figure}

In this section, we define understanding as a spectrum by presenting a series of levels. Understanding at each level is intended to be more difficult than the previous one, although they do not formally follow each other. Rather, they measure increasing levels of abstraction. To demonstrate the ideas, we also provide examples of questions that would be successfully answered by an entity that understands the Euclidean algorithm for GCD at each level. The full hierarchy is summarized in Figure \ref{fig:levels}. 

For simplicity, we present these levels as deterministic; however, they can be defined with a failure probability dependent on the entity's internal randomness and the required memory and time.

At the first level (denoted \textbf{Level 1}), the entity is capable of evaluating the algorithm on some `simple' examples, where the simplicity of an input is defined by the length of the execution path. At this level, the entity has some representation of the input-output mapping, whether or not it can formally express it. 

\begin{definition}[Level 1: Execution]
$\mathcal{E}$ understands $\mathcal{A}$ at Level 1 if there exists parameters $M_0, T_0$ such that the following holds: for any $x \in \Omega$ with $M(x) \leq M_0$ and $T(x) \leq T_0$, $\mathcal{A}_{\mathcal{E}}(x) = f(x)$.\\
\textbf{Example:} Compute GCD(24, 15). 
\end{definition}

At the next level, the entity can describe how it evaluates $f(x)$ in a language that it knows. Level 2 requires the entity to output the execution steps of the algorithm on $x$ as well as produce the correct answer. 

\begin{definition}[Level 2: Step-By-Step Evaluation]
$\mathcal{E}$ understands $\mathcal{A}$ at Level 2 if, given an $x\in \Omega$ with $M(x) \leq M_{\mathcal{E}}$ it can provide one of the following:

\begin{itemize}
\item the execution path in natural language or code
\item a flow chart or other unambiguous pictoral representation of the execution path
\end{itemize}
executed when running $\mathcal{A}$ on $x$.\\
\textbf{Example:} Compute GCD(462, 948) and show each step of the calculation.
\end{definition}

The next level will take this one step further, requiring the entity to produce a set of instructions that can be followed to produce the right answer for any input $x\in \Omega$.
\begin{definition}[Level 3: Representation]
$\mathcal{E}$ understands $\mathcal{A}$ at Level 3 if it understands at Levels 1 and 2, and it can produce one of the following:
\begin{itemize}
\item a formal representation; e.g., code for $\mathcal{A}$ in a Turing-complete programming language it knows, a structured natural language description, an abstract syntax tree or Turing machine diagram. 
\item an unambiguous description of the execution steps in natural language. 
\end{itemize}
\textbf{Example:} Write a function in a programming language you know that can compute the GCD of any two integers.
\end{definition}

The first three levels measure the ability of the entity to recall a procedure and execute a known set of instructions. We place these in the category of `shallow learning'; in Mayer's taxonomy, they fall under the cognitive processes of recognizing, recalling, and executing~\cite{mayer2002rote}. Note that all three levels could be achieved by a hard-coded script. 

The next two levels target deep learning, and measure cognitive processes in the `Understanding' and `Analyzing' categories. We split the next levels into two subtrees, to distinguish cognitive processes utilizing functional linguistic skills from those utilizing mathematical reasoning~\cite{mahowald2024dissociating}. 

At Level 4, the entity demonstrates an understanding of `why' the algorithm is constructed as it is. It requires them to provide an  example to illustrate a property (mathematical reasoning) or explain the existence of a property to a specified audience (reasoning with language).

\begin{definition}[Level 4a: Exemplification]
Given a property $P$ of an execution path of the algorithm $\mathcal{A}$, $\mathcal{E}$ can generate an $x\in \Omega$ which satisfies $P$ or report that none exists. 
\textbf{Example:} Give an integer $0<x<55$ that requires the greatest number of recursive steps to compute GCD$(55, x)$. Describe how you chose this number.\\
\end{definition}

\begin{definition}[Level 4b: Explanation]
Given $\mathcal{A}$, or a property $P$ satisfied by the execution path of $\mathcal{A}(x)$, and an audience $\mathcal{E}'$, the entity can produce a text in natural language that has the following characteristics:
\begin{itemize}
\item Accurately describes the steps of the algorithm/execution path. 
\item Abstracts or shortens the full description by referencing other algorithms known by the audience.
\item Uses examples and analogies to other algorithms known by the audience to convey intuition. 
\end{itemize}
\textbf{Example:} You are teaching a student who understands basic math operations but struggles with algebra and division with remainders. Explain how the Euclidean algorithm is used to find the greatest common divisor (GCD) of two given numbers, prioritizing intuition.
\end{definition}

At Level 5, the entity can reason on perturbations of the algorithm and perturbations of the input, and it can describe the effect on the execution path. Under reasoning with mathematics, this includes skills such as certifying if a modification to an algorithm changes the output for a subset of examples. Under reasoning with language, this includes describing the effects of modifying inputs, or answering counterfactual questions about modifications to the algorithm. 
\begin{definition}[Level 5a: Extrapolation]
The entity can answer questions about $\mathcal{A}$ of the following form.
\begin{itemize}
\item Given an algorithm $\mathcal{A'}$, the entity can determine whether $\mathcal{A}$ and $\mathcal{A'}$ produce the same output on all $x\in \Omega$. If not, it can find a counterexample such that $\mathcal{A}'(x) \neq \mathcal{A}(x)$.
\item Given a relation $R\subset\Omega \times \Omega$, the entity can find a pair $(x,x')\in R$ with different execution paths on $\mathcal{A}$.
\end{itemize}
%Given an algorithm $\mathcal{A'}$, the entity can determine whether $\mathcal{A}$ and $\mathcal{A'}$ produce the same output on all $x\in \Omega$. If not, it can find a counterexample such that $\mathcal{A}'(x) \neq f(x)$.\\
\textbf{Example: }Determine whether the following statement is true. If not, provide a counterexample. If $x>y$, then computing $GCD(2x,y)$ with the Euclidean algorithm requires more division operations than computing $GCD(x,y)$. 
\end{definition}
\begin{definition}[Level 5b: Counterfactual Reasoning] 
The entity can produce natural language descriptions of $\mathcal{A}$ of the following form. 
\begin{itemize}
\item Given an algorithm $\mathcal{A'}$ and an audience $\mathcal{E}'$, the entity can produce an explanation (c.f. Level 4b) contrasting the two algorithms.
\item Given a relation $R\subset\Omega \times \Omega$, the entity can describe a property highlighting the differences in execution paths for $(x,x') \in R$.
\end{itemize}

\textbf{Example: }Consider the Fibonacci sequence defined by $F(0) = 0$, $F(1) = 1$, and $F(n) = F(n-1) + F(n-2)$ for $n \ge 2$. Why do consecutive Fibonacci numbers result in the maximum number of iterations for the Euclidean algorithm?
\end{definition}

\section{Hypotheses}
We conducted an experiment on LLMs and human participants with two main goals; 1) to assess the proposed hierarchical scale (Figure \ref{fig:levels}) as a tool for comparing levels of understanding, and 2) to rate algorithm understanding across generations of GPT. We will assess the scale with a student survey, where we can use educational level as a basis for comparison. Then, we will apply the same questions to GPT and assess its understanding on the same scale. Related to these goals, we test the following hypotheses:\\
\textit{1. The understanding hierarchy (Figure \ref{fig:levels}) captures depth of understanding.}\\
We expect the fraction of correct answers to be non-increasing with higher levels of understanding. Furthermore, more education and training in algorithms should be reflected in the scores, so we expect graduate students to perform better than undergraduates.\\
\textit{2. Newer generations of GPT understand algorithms at a higher level than older generations.} \\
Concretely, we expect an increase in performance at higher levels between GPT-3.5 and GPT-4. \\
\textit{3. LLMs will exhibit a performance gap between natural language reasoning and mathematical reasoning tasks. }\\
We expect the difference in performance between these two types of tasks to be much smaller in students than LLMs. Furthermore, we expect that GPT may have a higher performance at Level 3, since GPT is fine-tuned on code generation and has been exposed to code for common algorithms.  

\section{Methods}

We use two classical algorithms to test our scale of understanding: the {\em Euclidean} algorithm for computing the greatest common divisor of two integers, and the {\em Ford-Fulkerson} algorithm for computing the maximum flow between two nodes on a directed graph with capacity constraints~\cite{ford1956maximal}. Both algorithms are widely taught in undergraduate computer science curricula. The Euclidean algorithm is relatively simple, and it is the first introduction to the concept of algorithm for many students. The Ford-Fulkerson algorithm, on the other hand, is more sophisticated and is typically introduced in an upper-level algorithms course. 

\subsection{Experimental Design} 
In this section, we describe how the assessment was constructed. This will serve as a guide to generalize the experiment to other algorithms. 

For each of the assessed algorithms, we produced a series of questions corresponding to each of the levels (Figure \ref{fig:levels}). 
\begin{enumerate}
\item A trivial instance of the problem. If the entity understands the input and output space, the problem can be answered without calculation.
\item An intermediate instance of the problem. This is answerable without a calculator for most undergraduates, but requires the entity to run some internal algorithm to compute each individual step. 
\item A coding problem. This problem requires the entity to translate a part of the algorithm to code.
\item Either an example problem (a) or an explanation problem (b). This question asks the entity to provide an example to illustrate a concept or explain the algorithm to a specified audience. 
\item Either a counter-example (a) or extension problem (b). This problem asks the entity to reason about modifications to the algorithm through calculation or explanation. 
\end{enumerate}

Examples of questions posed in the survey are included under the level definitions, and the full survey will be available in the Appendix. The most challenging part of constructing the assessment is in questions 4 and 5. An algorithm can be thought of as consisting of three parts - an input space, an output space, and a transformation procedure. The procedure can be further broken up into subroutines consisting of simpler algorithms that an entity may understand in other contexts. To construct an explanation problem, we specify an audience for the explanation, which cues the level of detail and the types of subroutines which can be referred to. We construct a counterfactual problem by modifying the input, output, or a key subroutine. 

\subsection{Human Survey}
We conducted a survey on students of algorithms courses at a premier CS-teaching university. Each student was assigned either the Euclidean or Ford-Fulkerson Algorithm at random, and was asked to rate their own understanding of the algorithm on a six-point scale.  Each survey consisted of five test questions to test their understanding. There were three versions of the survey for each algorithm, assigned at random. The questions are available in the Appendix. 

The number of participants in the survey was $n=34$ (10 doctoral and 24 undergraduate). Students who reported that they did not understand the algorithm or completed less than half of the survey questions were removed the analysis. This left $n=23$ students (10 doctoral and 13 undergraduate). Of these students, ten had some teaching assistant experience in algorithms classes. 

\subsection{LLM Experiments} Several versions of ChatGPT were presented with the same surveys given to the human participants; each survey was started in a fresh chat session, and within the survey, previous questions and responses were included in the chat history. We also included a system prompt to prime GPT and encourage conciseness in the responses.  

GPT was also queried using randomized versions of the survey. For evaluation questions, the input values were assigned uniformly at random within a given range. For the flow questions, the graph structure was also varied slightly. For the code questions, we took an example code implementation of the algorithm, masked a key section, and asked GPT to fill in the missing part. We also included several versions of the example, explanation, and extension questions. 

%Three versions of GPT were tested: GPT-3.5 Turbo, GPT-4, and GPT-4o. 

\subsection{Evaluation}
\label{sec:eval}
Each question is rated on a scale from zero to two. With the exception of the explanation questions, the scores have the following interpretations: (0) incorrect; (1) partially correct, surface level; (2) completely correct, thorough.
\subsubsection{Evaluating Explanations}
The quality of explanations and summaries can be subjective; however, they offer a deep insight into the subject's understanding of the material. We evaluate the explanations on three axes.
\begin{enumerate}
\item Correctness; the explanation is accurate and includes the key ideas of the algorithm.
\item Audience adaptation; the explanation is tuned to the audience, and the level of detail matches their prior knowledge.
\item Intuitiveness; the explanation conveys intuition; via contrast, example, analogy etc. and uses clear language. 
\end{enumerate}
Summaries and explanations are by definition \textit{selective} and not necessarily complete~\cite{mittelstadt2019explaining}. The ability to identify key ideas is part of what differentiates explanation (Level 4) from the production of instructions (Level 3). An explanation is awarded 2/3 of a point for each bullet, for a maximum score of 2 per question. 

\section{Results}

\begin{figure}[h]
 \centering
    \includegraphics[width = \columnwidth]{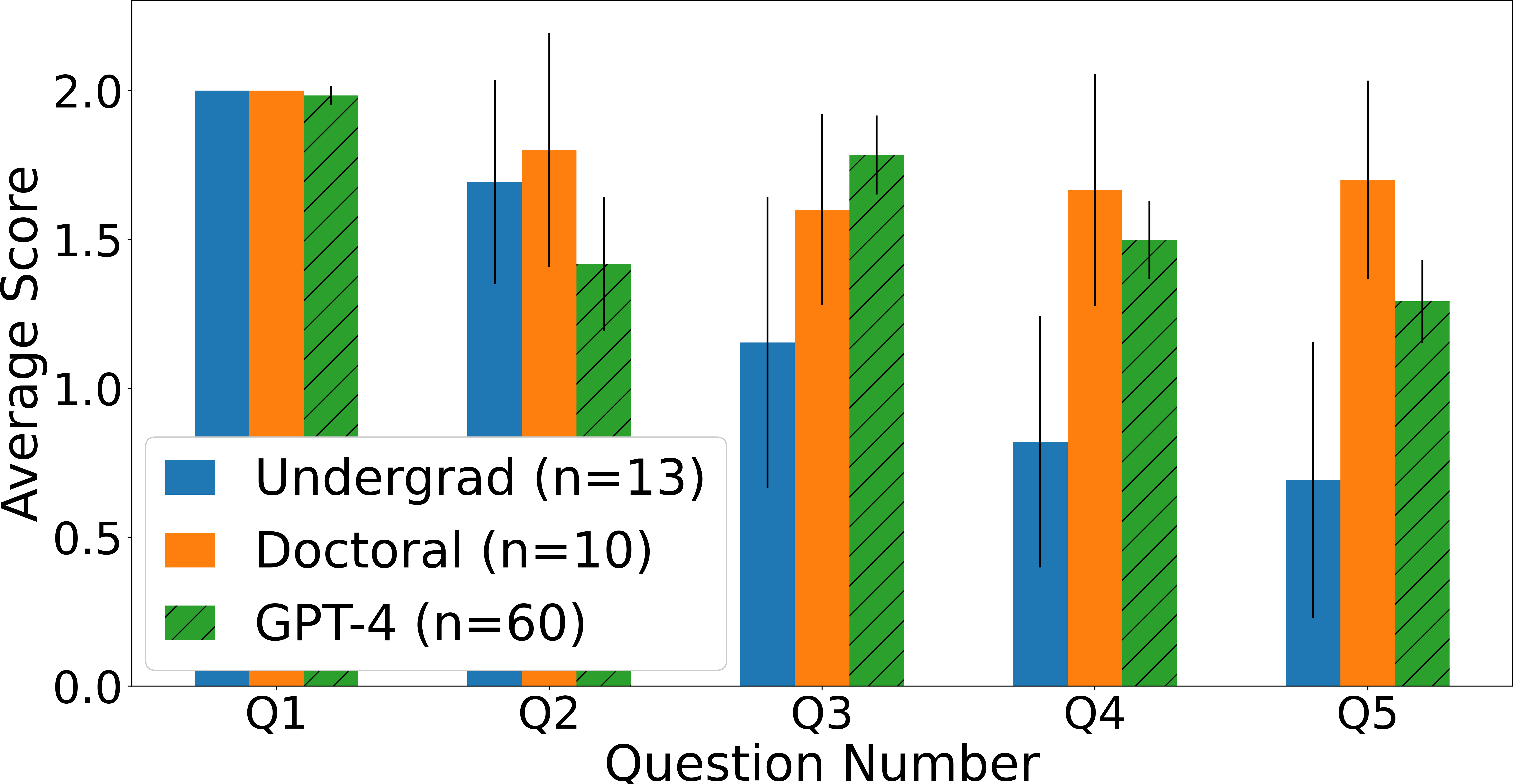}
	\caption{The average scores across students who self-reported that they understood the algorithm. Number of records is $n=13$ (undergraduate) and $n=10$ (graduate) respectively. The average scores for GPT-4 are across 60 randomized versions of the surveys. Error bars are 95\% confidence intervals.}
	\label{fig:mean_score}
\end{figure}

\begin{figure}[h]
 \centering
    \includegraphics[width = \columnwidth]{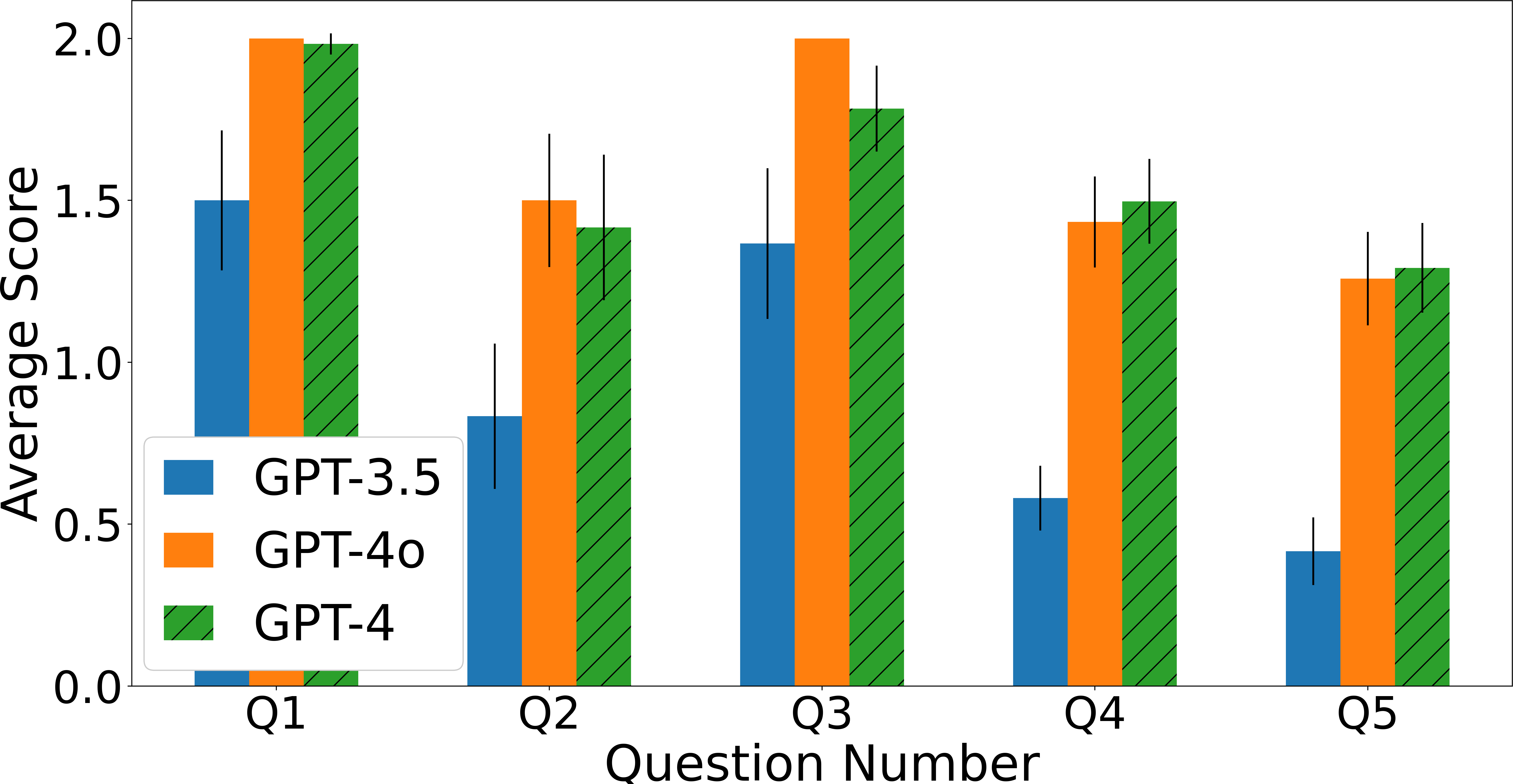}
	\caption{The average score between three versions of GPT, across 30 random surveys for each of GCD and Max Flow. Error bars show the 95\% confidence interval. }
	\label{fig:mean_score_gpt}
\end{figure}
\begin{figure}[h]
 \centering
    \includegraphics[width = \columnwidth]{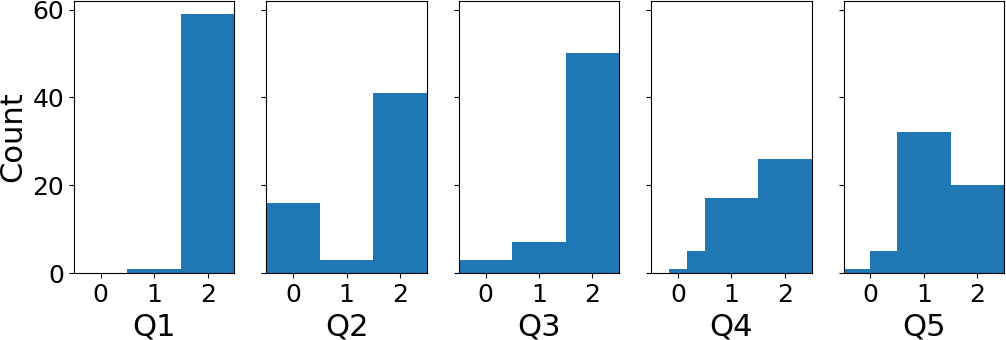}
	\caption{The distribution of scores per question for GPT-4.}
	\label{fig:perq}
\end{figure}

\paragraph{Hypothesis 1} \textit{The understanding hierarchy (Figure \ref{fig:levels}) captures depth of understanding.} 

Overall, 85\% of students indicated they understood the algorithm, with most students reporting that they ``know the algorithm and have a fair understanding of it''. For further analysis, we only consider students who stated that they understood the algorithm. 
We compare undergraduate and doctoral levels in Fig. \ref{fig:mean_score}. Across all students, the accuracy on the questions was highest for Question 1, and decreased uniformly through Question 5. 
Doctoral students performed better on average than undergraduates ($p< 0.05$). They received higher scores on Q4 and Q5 ($p < 0.05$), while the differences on Q1, Q2, and Q3 were not statistically significant. 

\begin{figure}
        \centering
        \includegraphics[width=\columnwidth]{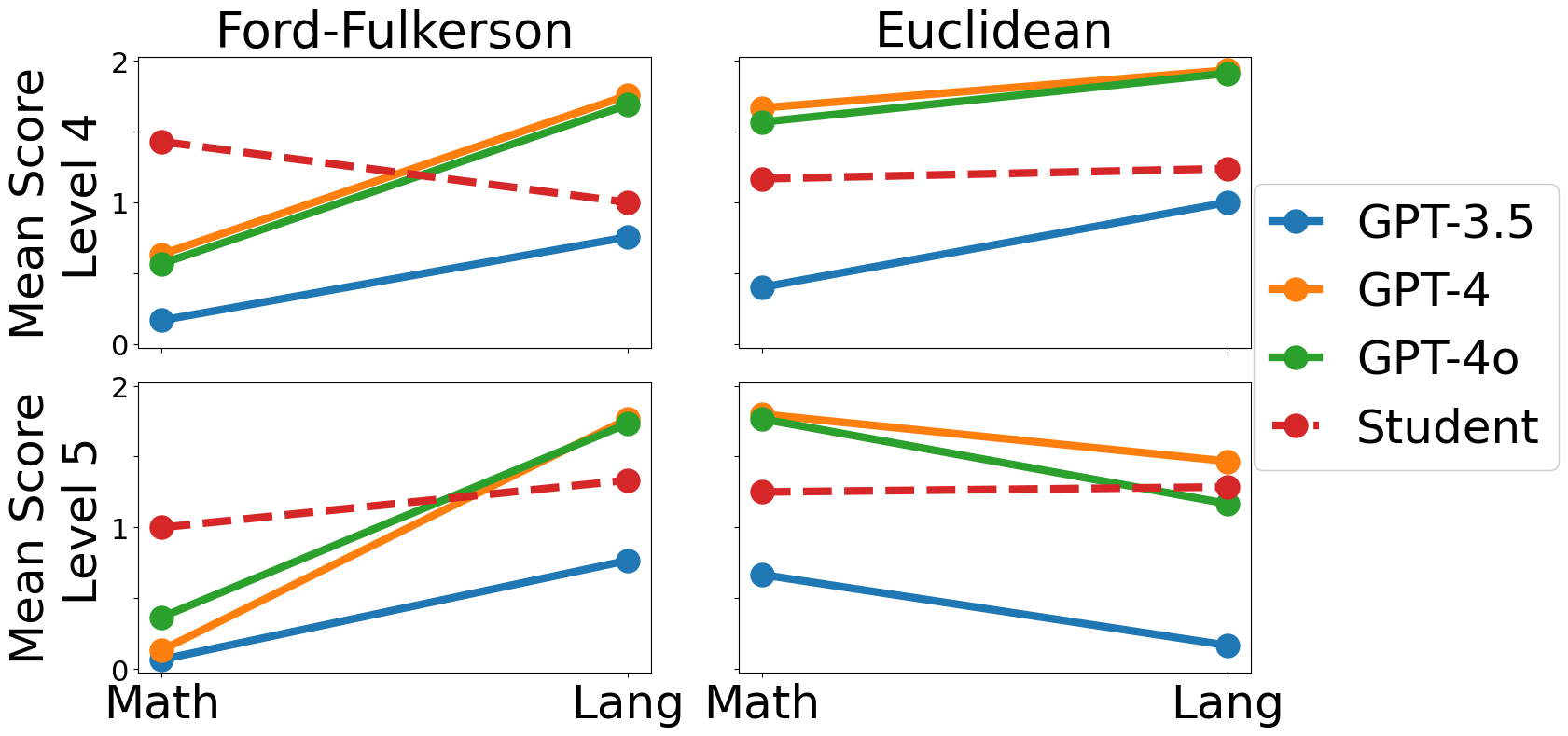}
        \caption{The difference in mean performance between mathematical and natural language reasoning tasks on Ford-Fulkerson (Left) and the Euclidean algorithm (right). The top graphs show tasks at Level 4, while the bottom graphs show tasks at Level 5}
        \label{fig:hyp3_compare}
\end{figure}
\paragraph{Hypothesis 2}\textit{Newer generations of GPT understand algorithms at a higher level than older generations.} \\
Among versions of GPT, GPT-4 and GPT-4o performed about the same, and the differences in their overall scores were not significant (Figure \ref{fig:mean_score_gpt}). Both GPT-4 and GPT-4o demonstrated an increase in score on every question compared to GPT-3.5 ($p<0.05$). 

The response score of GPT-4 was close to that of graduate students, as shown in Figure \ref{fig:mean_score}. Doctoral students scored better than GPT-4 on the extension questions (Q5) to a statistically significant degree. LLMs on average out-performed the undergraduate students on questions 3, 4, and 5 ($p < 0.05$), while the differences on Q1 and Q2 were not statistically significant. 
\paragraph{Hypothesis 3}\textit{LLMs will exhibit a performance gap between natural language reasoning and mathematical reasoning tasks. }\\
As shown in Figure \ref{fig:hyp3_compare}, all three versions tested performed better on language tasks than on mathematical reasoning tasks for Ford-Fulkerson (significant with $p< 0.05$) despite student performance being the same or slightly worse. For GCD, the versions performed better on language tasks than on mathematical reasoning tasks on Level 4, but the same or slightly worse on Level 5. 

We also hypothesized that the performance on code tasks would be higher compared to the performance on evaluation and reasoning tasks. We find that this does hold. As shown in Figure \ref{fig:mean_score_gpt}, LLM performed better on the coding tasks (Q3) than on the evaluation tasks (Q2), while the students exhibited the opposite trend (Figure \ref{fig:mean_score}). 
\paragraph{Prompting with examples.} We also investigated whether the use of example responses can improve LLM responses to Max Flow problems.  Each problem was introduced with the following prompt, priming the use of chain of thought reasoning: ``Compute the maximum flow between A and $\langle$Sink Vertex$\rangle$. List each augmenting path and the flow along the path at each step.'' Then the graph was described as a list of edges and capacities. We tested 200 randomly instantiated maxflow problems (100 trivial and 100 intermediate), with and without a correct example response included in chat history.

\begin{figure}[h]
 \centering
    \includegraphics[width = \columnwidth]{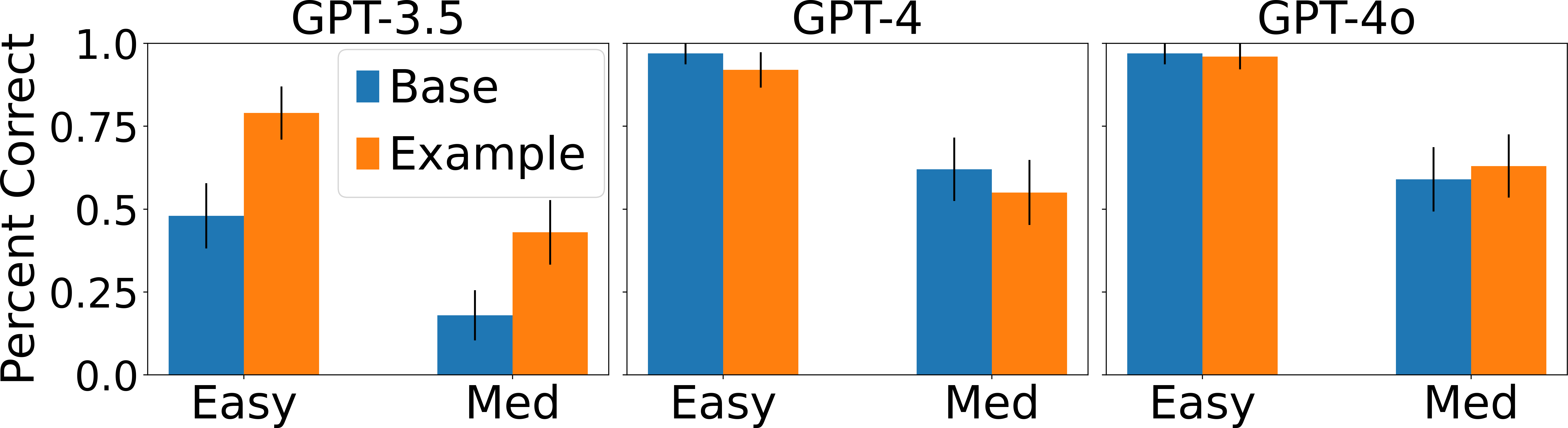}
	\caption{Accuracy of GPT versions on 100 (randomized) trivial/intermediate max flow problems. Base accuracy is in blue and accuracy when prompted with a correct example is in orange. Error bars are 95\% confidence intervals.}
	\label{fig:gpt-examples}
\end{figure}

We find that this strategy generally causes the response to mimic the structure of the example response. As shown in Figure \ref{fig:gpt-examples}, including an example response improves accuracy for GPT-3.5, but has little to no effect on GPT-4o, and marginally decreases the accuracy for GPT-4. One possible explanation for this phenomenon is that GPT-4 naturally responds to the prompt with effective chain of thought reasoning; therefore, instructing it to reason in a specific format does not improve its reasoning abilities, and may in fact interfere with them.

\paragraph{Qualitative Comparison.} 
\textit{Generating Examples.} As shown in Figure \ref{fig:hyp3_compare} (left), all versions of GPT struggled to produce an example of a graph satisfying a prescribed property. This could be attributed to GPT's known difficulties with mathematical calculation --- after all, in order to know if the execution of the algorithm satisfies a property, it might have to first execute the algorithm, where GPT tends to make errors even with chain-of-thought prompting~\cite{wei2022chain}. However, our evidence suggests a deeper issue with categorizing inputs. 

As an example, we asked the following question: ``Give an example of a graph where the Ford-Fulkerson algorithm computes exactly six augmenting flows before terminating. Write the example as a list of edges and capacities.'' This question is deliberately imprecise, and the intended answer is a graph with a source $s$, a sink $t$, and six intermediate vertices connected to both, leading to six parallel paths from s to t. However, more complex graphs could also be correct.  

In the human survey, out of nine students, six described the graph made of parallel paths (the other three did not attempt the question). It is reasonable to expect that these students did not mentally execute the Ford-Fulkerson algorithm in order to verify their answer. Instead, they likely had some knowledge of the concept of parallel paths, and were able to leverage this knowledge to retrieve an example. 

The graphs given by GPT did not follow any discernible pattern. Over all trials, none required six augmenting paths (all examples reviewed were too small, admitting at most 5 augmenting paths regardless of the path-finding algorithm). We argue that this suggests that GPT lacks an ability to manipulate its representation of the input space to produce useful shortcuts. \\
\textit{Hedged Responses.} Throughout GPT's responses, we found frequent instances of GPT incorrectly `hedging' its answers. When stating a true property of an algorithm, it frequently includes qualifiers such as `usually' or `potentially' that make the statement incorrect. E.g., responding to a question about the relationship between the Euclidean algorithm and the Fibonacci sequence, GPT-4 included the following line (emphasis ours):
\begin{quote}
\dots Each Fibonacci number is the sum of the two preceding ones, with the sequence beginning as \( F(0) = 0 \), \( F(1) = 1 \), \( F(2) = 1 \), \( F(3) = 2 \), and so forth. This means \textit{each number in the sequence is relatively close to the sum of the two preceding numbers}.
\end{quote}
In another response, it stated that the golden ration $\phi$ was ``one of the most irrational numbers". In several responses, it stated that the Ford-Fulkerson algorithm `potentially' tracks the capacity of reverse edges (not tracking would simply be an incorrect implementation of the algorithm). While this writing style may be an asset for subjects such as politics or health where being overly confident may cause harm, it causes statements about objective mathematical properties to be incorrect. This highlights the need for caution when using GPT-4 for teaching. \\
\textit{Hallucinations.} All three versions of GPT occasionally produced hallucinations in the responses. The most common type of hallucination occurred when it tried to produce counter-examples. When asked to evaluate the (true) statement `GCD(a,b,c) = GCD(GCD(a,b), GCD(b,c)), GPT often tried to disprove it with a counter-example. GPT-3.5 generally claimed that its counter-example disproved the statement, despite the fact that the two sides of the equation were evidently the same.
\begin{quote}
The statement is false. Counterexample: Let a = 8, b = 12, c = 6. gcd(8, 12) = 4, gcd(12, 6) = 6. gcd(8, 12, 6) = 2, which is not equal to gcd(4, 6) = 2.
\end{quote}
GPT-4 and GPT-4o both recognized when the `counter-examples' failed to disprove the original statement. However, both continued to attempt to present counter-examples - in some cases, after many failed attempts, the responses became nonsensical. For example, from GPT-4:
\begin{quote}
\dots So GCD(GCD(1, 14))= 1). Therefore: GCD(6, 35, 14) = 1 And that verifies the consistency. Given the importance of a concept, premise restated true in a broader context, counter intuitive aligning confirm example specifics prove legitimacy \dots
\end{quote}
Neither type of hallucination is observed in humans. 
\section{Discussion} We have presented a hierarchical scale for quantifying the understanding of algorithms. We verified its predictions empirically on human subjects and used it to compare generations of GPT and students. Our results show a significant improvement from GPT-3.5 to GPT-4/4o at all levels of algorithm understanding. We find that GPT-4 possesses a functional understanding of both the Euclidean algorithm and the Ford-Fulkerson algorithm, and it performs on a comparable level with CS PhD students. However, its reasoning abilities with mathematics lag behind its reasoning with language, and its understanding is not fully robust.

All versions of GPT were nearly perfect on code generation tasks. This is in line with other findings showing the competence of GPT in code generation~\cite{vaithilingam2022expectation, savelka2023thrilled}. This trend was not observed in the student respondents, who performed better at evaluating the algorithm than producing code on average. One possible reason for this is that the code for common algorithms, such as those tested, are prevalent in GPT's training data. Code is highly structured, so even if the particular implementation of the algorithm has not been observed by GPT, it could replicate the changes, for example in variable names, by statistical inference.  

Another trend is that GPT generally performed better on reasoning with language than with mathematics, while student performance was about the same. This difference goes beyond algebraic computations - GPT struggles with questions testing common-sense graph reasoning that humans can answer easily. However, for explanation questions and reasoning questions that do not involve examples, GPT-4 and 4o give consistently quality responses. This may suggest that mathematical examples play a larger role in human understanding than in LLMs.

This result begs the question: is GPT actually reasoning, or can the responses be explained by more superficial statistical correlations? If many similar questions and answers can be found in its training data, then GPT may be able to produce correct answers by leveraging statistical correlations between the input and the correct response. Can this really be called {\em understanding}? To this we make two points. First, the evaluation questions are instantiated with random values, and the exact questions are almost certainly novel. We argue that this suggests that GPT must be making some nontrivial transformation to produce correct answers. 

Second, in order to determine whether GPT is reasoning, the concept of reasoning itself needs to be precisely defined. Our study indicates that GPT-4 has a sophisticated internal representation, e.g., its representation of the Euclidean algorithm includes its relationship to Linear Diophantine equations and mathematical properties of GCD. It is able to retrieve these properties under a variety of contexts. In a similar vein, it is likely that a student answering these questions has also been exposed to these properties. When evaluating the usefulness of a representation, we do not necessarily need to account for how the representation was created. 

\paragraph{Limitations.} Our results show that the hierarchy of understanding is consistent with classical notions of depth of understanding when tested on humans. While the results are also consistent with later versions of GPT having a `better' understanding of the tested algorithms than undergraduates, such a conclusion does not follow. We worked with a limited population size, and the difference is confounded by other factors such as subject fatigue. Further research is needed to compare the quality of GPT and human responses to questions about algorithms. Despite these limitations, we feel that our scale makes progress towards a testable definition of understanding and can be extended to other algorithmic and similarly precise realms of understanding.
\section*{Acknowledgements}
The authors are grateful to Rosa Arriaga, Adam Kalai and Sashank Varma for helpful discussions. This work was funded in part by NSF Award CCF-2106444 and a Simons Investigator award. 
\bibliography{references}
\newpage
.
\newpage
\appendix
\section{Survey Questions}

\begin{figure}[h]
 \centering
 \begin{subfigure}[t]{\columnwidth}
    \includegraphics[width = 0.9\columnwidth]{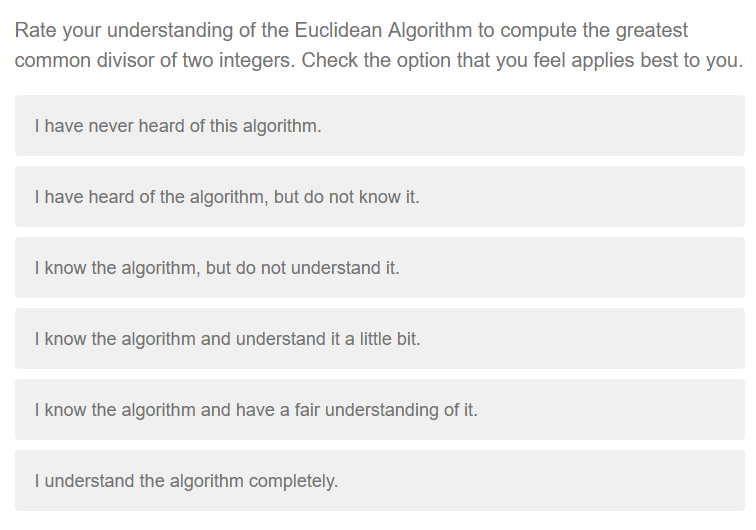}
\end{subfigure}
\begin{subfigure}[t]{\columnwidth}
\includegraphics[width = 0.9\columnwidth]{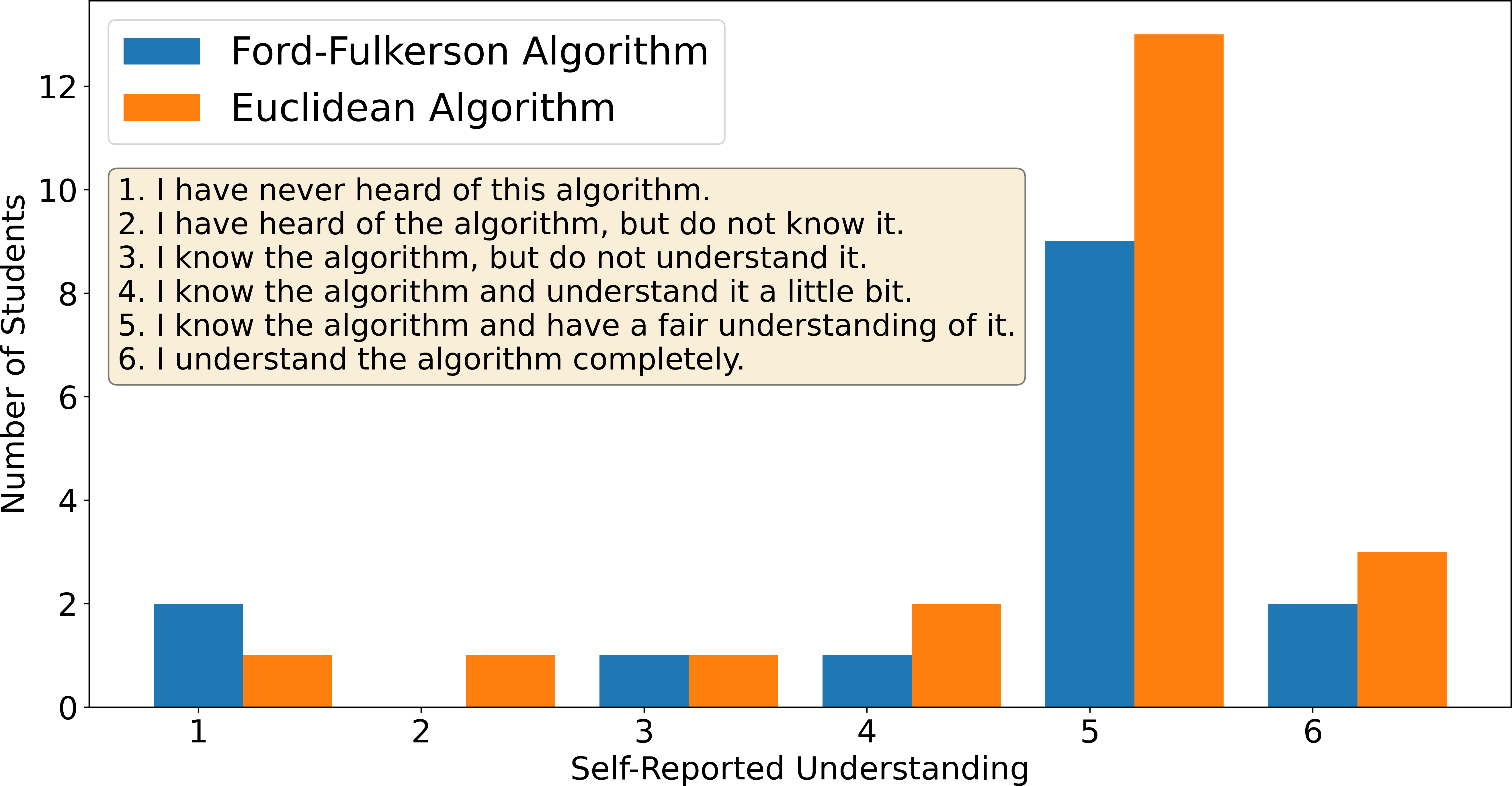}
 \end{subfigure}
 \caption{The six point-scale shown to students to self-report their understanding, and the distribution of responses.}

	\label{fig:self-report}
\end{figure}

This section contains the text of the survey given to human subjects. Each respondent was asked some preliminary questions about their course level and experience with algorithms. They were assigned either the Ford-Fulkerson algorithm or the Euclidean algorithm at random and asked to self-report their understanding (Figure \ref{fig:self-report}). Some students who reported that they did not understand the Ford-Fulkerson algorithm were then given the survey for the Euclidean algorithm.

The student was then randomly assigned one of three survey versions, shown below.
\subsection{Euclidean Algorithm}
\subsubsection{Version 1}
\begin{enumerate}
\item Compute GCD(24, 15), and show each step of the algorithm.
\item Compute GCD(462, 946), and show each step of the algorithm.
\item Using a programming language you are familiar with, write code for a function gcd(a,b), which computes the greatest common divisor of two integers a and b using the Euclidean algorithm.
\item You are instructing a student in an algebra course. The student is familiar with basic mathematical operations such as addition, subtraction, multiplication and division. However, she struggles with algebraic equations and division with remainders. Explain the Euclidean algorithm to her, prioritizing conveying intuition.
\item Suppose gcd(a,b) = x and gcd(b,c) = y. Does gcd(a,b,c) = gcd(x,y)? Explain your reasoning.
\end{enumerate}
\subsubsection{Version 2}
\begin{enumerate}
\item Compute GCD(24, 15), and show each step of the algorithm.
\item Compute GCD(4088, 1241), and show each step of the algorithm.
\item Using a programming language you are familiar with, write code for a function gcd(a,b), which computes the greatest common divisor of two integers a and b using the Euclidean algorithm
\item Consider computing GCD(55, x) for an input x. Give an integer $0<x<55$ that requires the greatest number of recursive steps to compute GCD(55, x). Describe how you chose this number.
\item Determine whether the following statement is true. If not, provide a counterexample. 
For any two positive, nonzero integers a,b, the equation sa + tb = GCD(a,b) has exactly one solution where s and t are integers.

\end{enumerate}
\subsubsection{Version 3}
\begin{enumerate}
\item Compute GCD(24, 15), and show each step of the algorithm.
\item Compute GCD(1008, 468), and show each step of the algorithm.
\item Using a programming language you are familiar with, write code for a function gcd(a,b), which computes the greatest common divisor of two integers a and b using the Euclidean algorithm.
\item You are speaking with a mathematics student who understands modular arithmetic. Explain the proof that the Euclidean algorithm finds the greatest common divisor of two numbers.
\item The Fibonacci numbers are a sequence F(n) where F(0) = 0, F(1) = 1, and F(n) = F(n-1) + F(n-2) for any $n\geq 2$. Consecutive Fibonacci numbers can be thought of as `worst case' inputs for the Euclidean algorithm. Can you explain why?
\end{enumerate}
\subsection{Ford-Fulkerson Algorithm}
\subsubsection{Version 1}
\begin{enumerate}
\item Compute the maximum flow between A and B.  List each augmenting path and the flow along the path at each step.
\begin{figure}[H]
 \centering
    \includegraphics[width = 0.9\columnwidth]{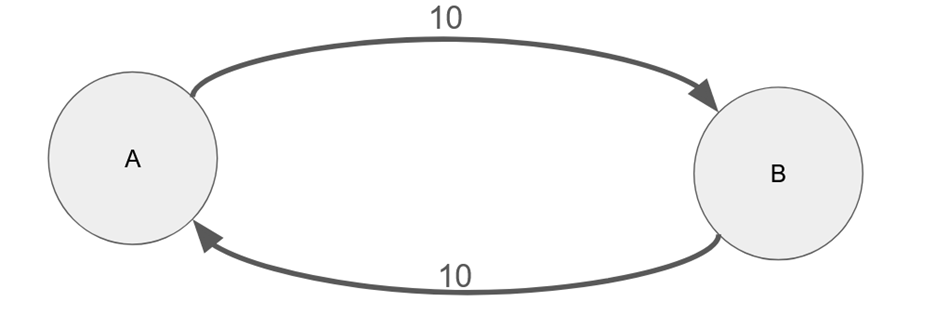}
\end{figure}
\item Compute the maximum flow between A and D.  List each augmenting path and the flow along the path at each step.
\begin{figure}[H]
 \centering
    \includegraphics[width = 0.9\columnwidth]{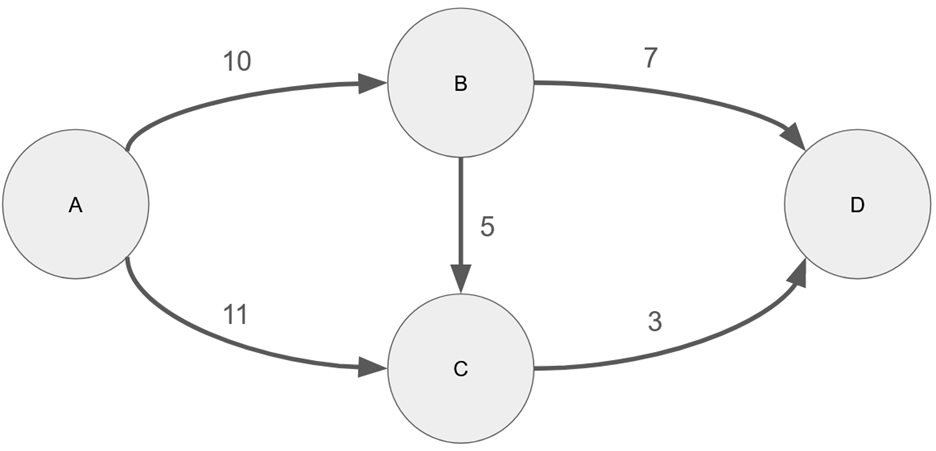}
\end{figure}
\item The following is an implementation of breadth-first search in Python with one line missing. Write a condition to replace the highlighted text. If you are not familiar with Python syntax, you may use your best guess.
\begin{figure}[H]
 \centering
    \includegraphics[width = 0.9\columnwidth]{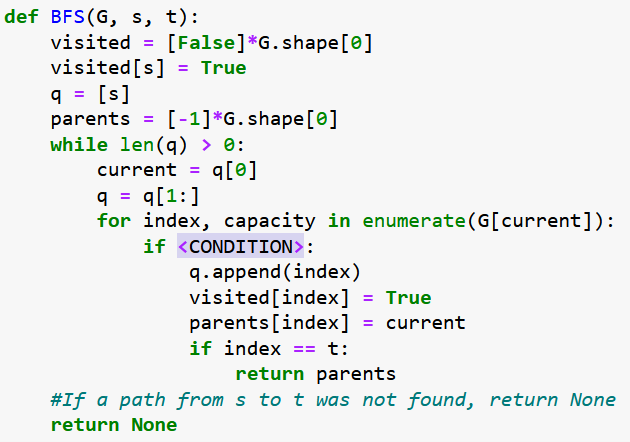}
\end{figure}
\item Give an example of a graph where the Ford-Fulkerson algorithm computes exactly six augmenting flows before terminating. Write the example as a list of edges and capacities.
\item Consider an implementation of Ford-Fulkerson which uses a breadth-first search to find the augmenting path. Give an example which illustrates why the algorithm needs to track residual capacity of reverse edges.
\end{enumerate}
\subsubsection{Version 2}
\begin{enumerate}
\item Compute the maximum flow between A and B.  List each augmenting path and the flow along the path at each step.
\begin{figure}[H]
 \centering
    \includegraphics[width = 0.9\columnwidth]{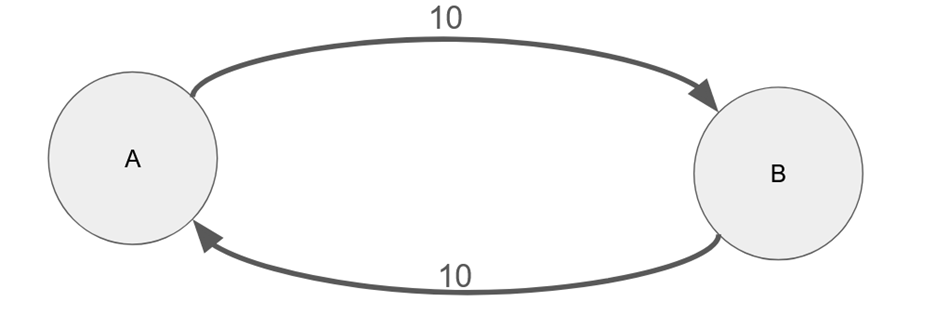}
\end{figure}
\item Compute the maximum flow between A and D.  List each augmenting path and the flow along the path at each step.
\begin{figure}[H]
 \centering
    \includegraphics[width = 0.9\columnwidth]{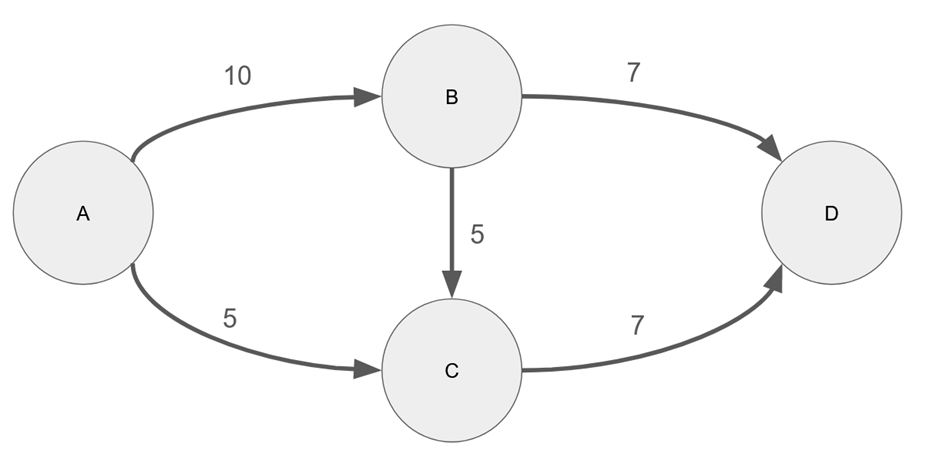}
\end{figure}
\item The following is an implementation of Ford-Fulkerson in Python with two lines missing. Write code to replace the highlighted text. If you are not familiar with Python syntax, you may use your best guess.
\begin{figure}[H]
 \centering
    \includegraphics[width = 0.9\columnwidth]{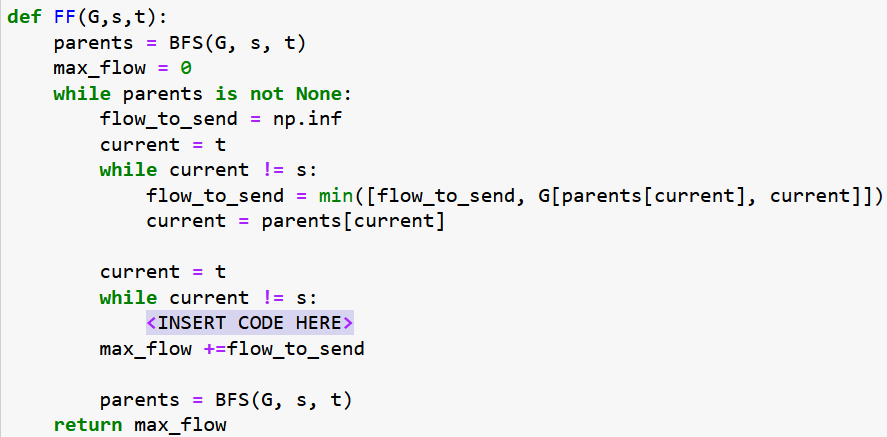}
\end{figure}
\item You are a civil engineer at a city planning commission. Your boss has asked you whether it is possible to calculate the maximum volume of traffic that can be routed between two destinations in the city.  Describe the Ford-Fulkerson algorithm, but since he is a busy man, do not bore him with the details.
\item Suppose you are an event planner, and you need to determine the maximum amount of traffic that can be routed from the airport to one of two event venues. In other words, you want to compute the maximum flow between a source s and two sinks t1 and t2. How would you implement this?
\end{enumerate}
\subsubsection{Version 3}
\begin{enumerate}
\item Compute the maximum flow between A and B.  List each augmenting path and the flow along the path at each step.
\begin{figure}[H]
 \centering
    \includegraphics[width = 0.9\columnwidth]{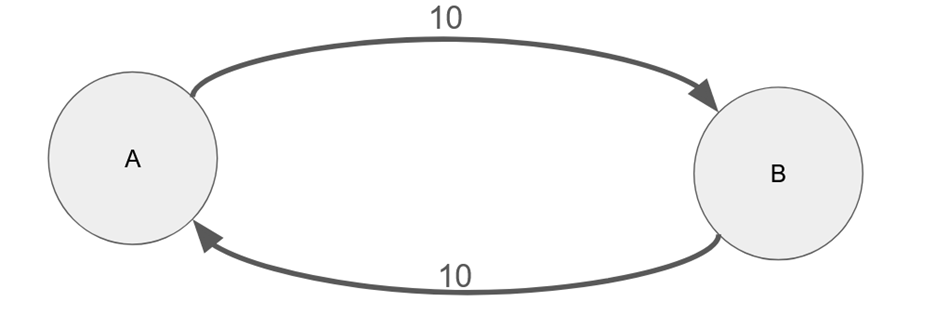}
\end{figure}
\item Compute the maximum flow between A and D.  List each augmenting path and the flow along the path at each step.
\begin{figure}[H]
 \centering
    \includegraphics[width = 0.9\columnwidth]{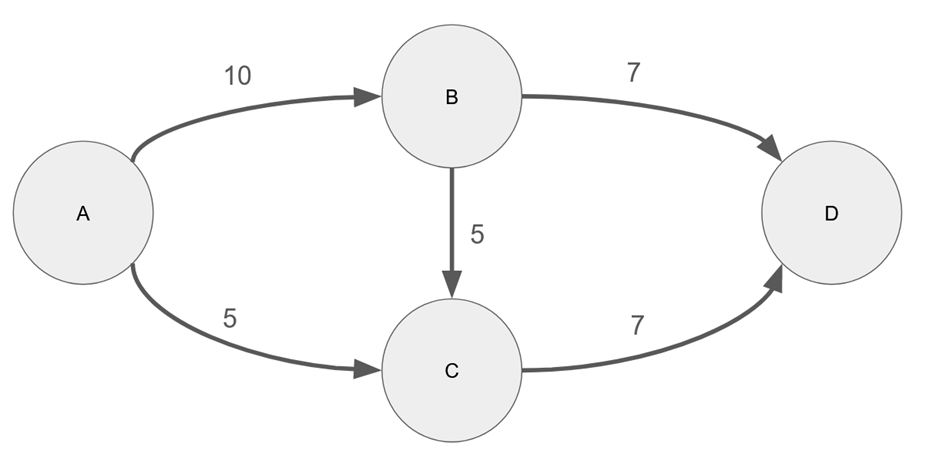}
\end{figure}
\item The following is an implementation of Ford-Fulkerson in Python with three lines missing. Write code to replace the highlighted text. If you are not familiar with Python syntax, you may use your best guess.
\begin{figure}[H]
 \centering
    \includegraphics[width = 0.9\columnwidth]{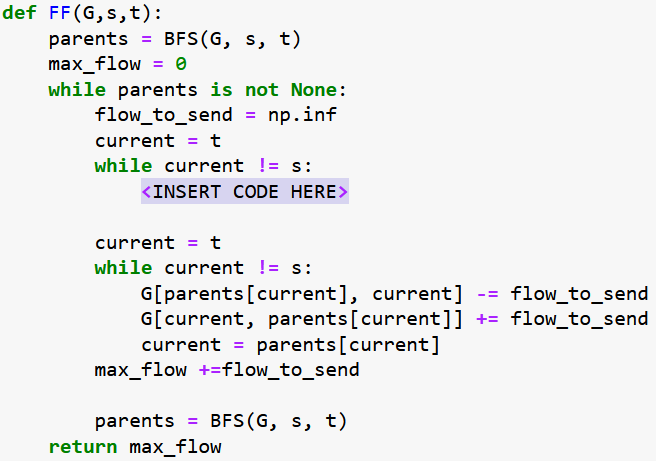}
\end{figure}
\item You are assisting a student in an algorithms course. He is struggling to understand the Ford-Fulkerson algorithm for computing MAXFLOW. Describe it to him, prioritizing conveying intuition.
\item An s-t cut is a set of edges which, when removed, divide vertex s and vertex t into separate components. For example, the dotted edges in the graph below are an A-F cut with capacity 24. \begin{figure}[H]
 \centering
    \includegraphics[width = 0.9\columnwidth]{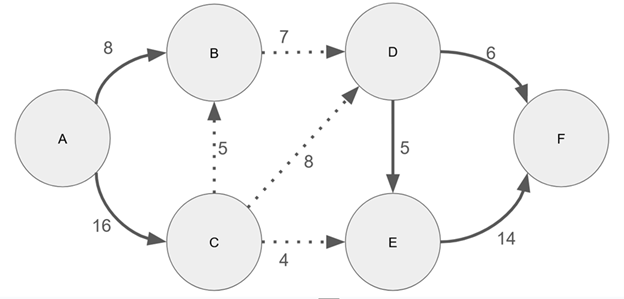}
\end{figure}

It turns out that the maximum s-t flow on a graph with capacities is equal to the minimum capacity of any s-t cut. Describe your intuition for why this might be true.

\end{enumerate}
\section{Further details on LLM experiments}
In each experiment on GPT, the query included a system prompt to encourage the model to produce only relevant information. The system prompts used in the experiments are as follows: 
\begin{quote}
`You will answer a series of questions related to computing the maximum flow on a directed graph. You provide concise responses and do not include detail or explanations unless explicitly requested by the user.'\\
`You will answer a series of questions related to the Euclidean algorithm for computing the greatest common divisor (gcd) of two integers. You provide concise responses and do not include detail or explanations unless explicitly requested by the user.'
\end{quote}

GPT was queried via the OpenAI Chat Completions API using the default parameters (e.g. temperature=1). Specific versions queried are `gpt-3.5-turbo-0125',`gpt-4-turbo-2024-04-09', and `gpt-4o-2024-05-13'. 

We conducted two experiments with GPT. In the first, GPT was asked randomized versions of the questions given to students (a {\em quiz}), with the images being replaced with text descriptions of the graphs. For each quiz, the questions corresponding to each of the levels of understanding were asked in order, and the previous questions and responses were included in the API query. This quiz was repeated thirty times for each GPT version and algorithm. All trials were hand-graded on a scale from zero to two. 

In the second experiment, GPT was queried with only the simple and intermediate evaluation questions for Max Flow (with varying graph structure and random weights). In one trial, GPT was first asked the simple question followed by the intermediate (with its response to the simple question passed to the API in the chat history). In the second trial, a sample response to an intermediate evaluation was passed to the chat history. The responses were checked by hand for correctness, and the rate of correctness was recorded. 

\section{Additional Results}
\begin{figure}[hb]
 \centering
    \includegraphics[width = \columnwidth]{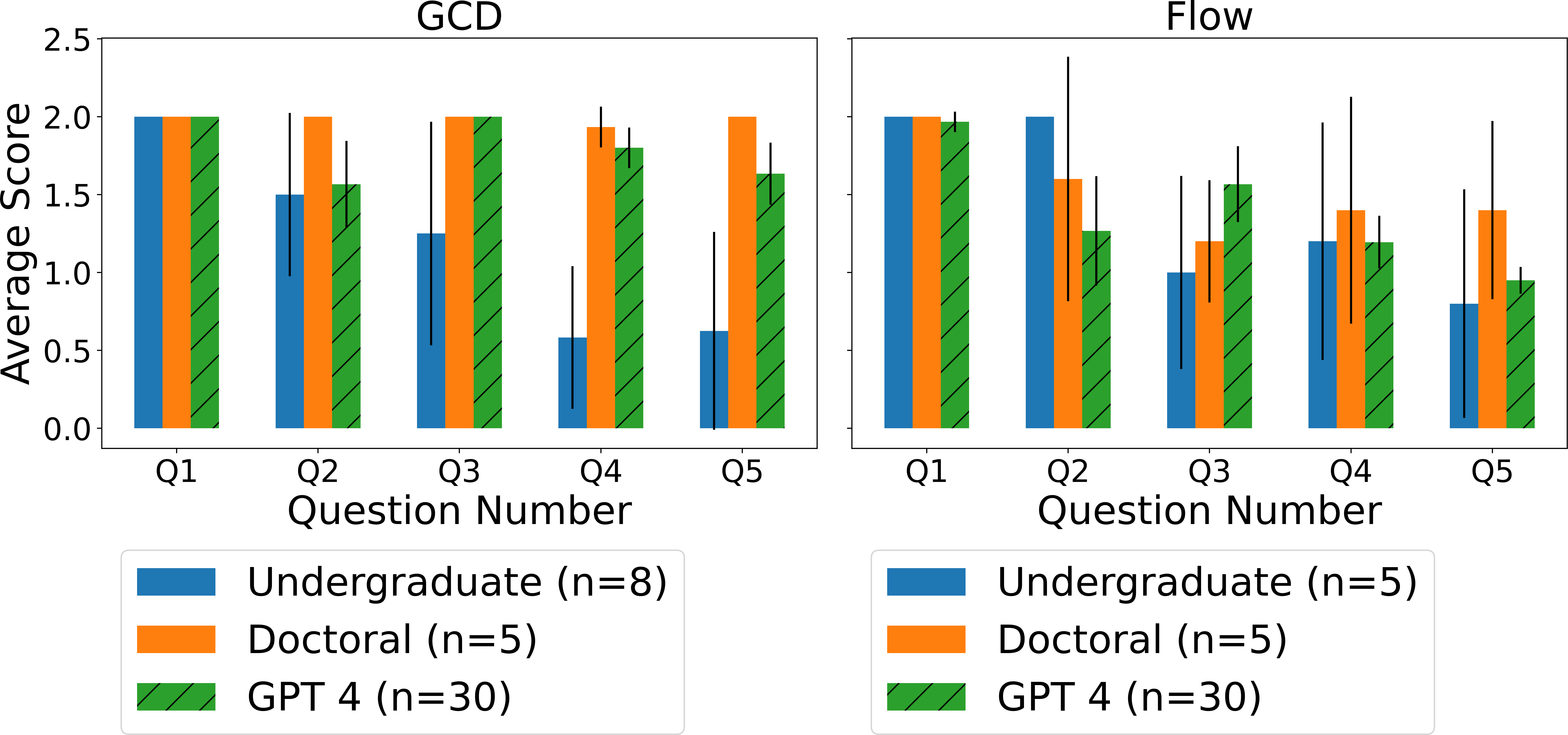}
	\caption{The average scores across students who self-reported that they understood the algorithm on the Euclidean algorithm (left) and Ford-Fulkerson algorithm (right). Error bars show the 95\% confidence interval.}
	\label{fig:mean_scores_by_alg}
\end{figure}

\begin{figure}[ht]
 \centering
    \includegraphics[width = \columnwidth]{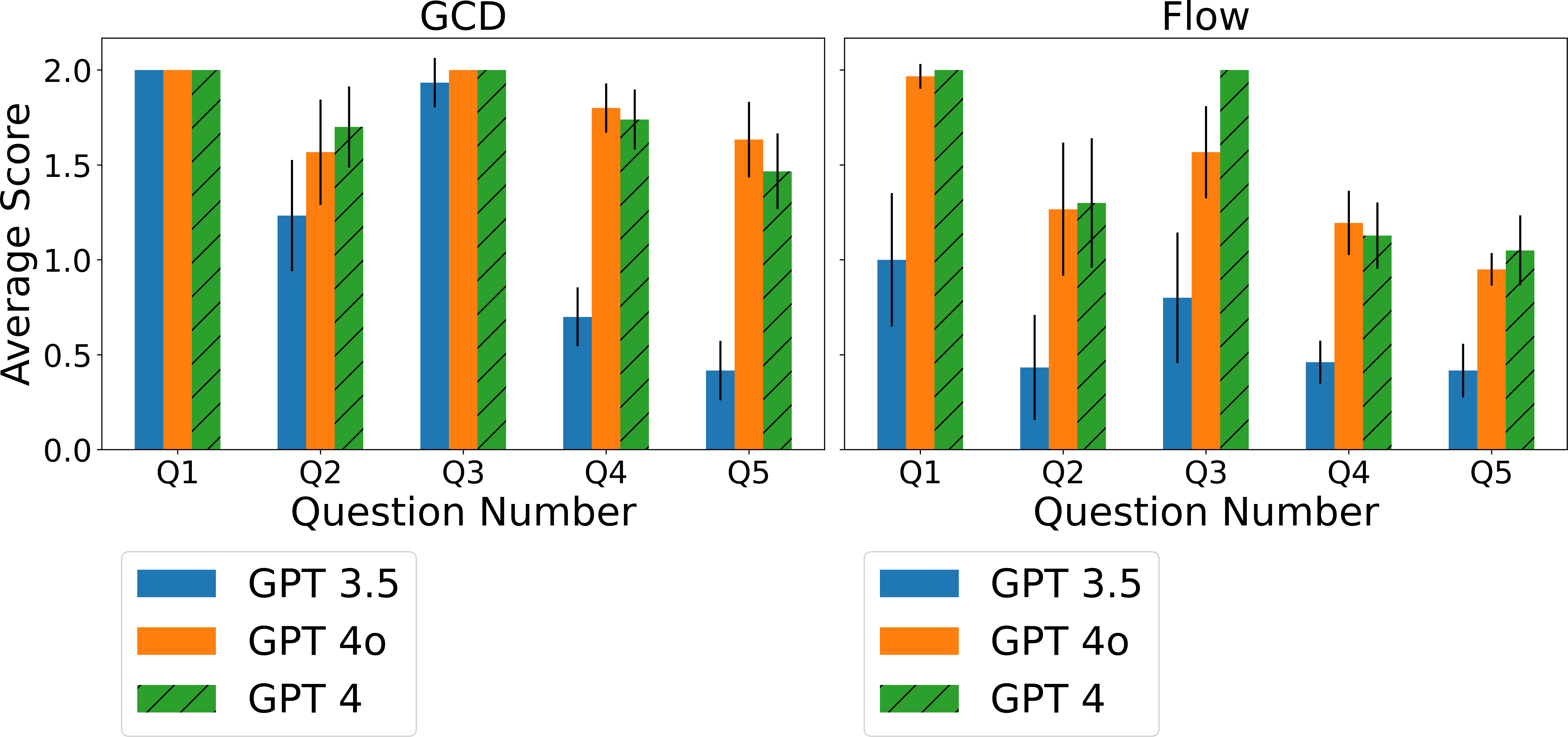}
	\caption{The average scores between versions of GPT on the Euclidean algorithm (left) and Ford-Fulkerson algorithm (right). Error bars show the 95\% confidence interval.}
	\label{fig:mean_scores_by_alg_gpt}
\end{figure}

Figures \ref{fig:mean_scores_by_alg} and \ref{fig:mean_scores_by_alg_gpt} show the mean score per question number (originally presented in Figures \ref{fig:mean_score} and \ref{fig:mean_score_gpt}) split by algorithm.


In Table \ref{tab:meanscore}, we include the mean performance for each of the seven types of questions in the five studied populations (Graduate, Undergraduate, and the three GPT versions).

\begin{figure}

\begin{tabular}{|l|rrrrr|}
\hline
\bfseries GCD & GPT 3.5 & GPT 4 & GPT 4o & UG & Grad \\
\hline
Q1 & \bfseries 2.00 & \bfseries 2.00 & \bfseries 2.00 & \bfseries 2.00 & \bfseries 2.00 \\
Q2 & 1.23 & 1.57 & 1.70 & 1.50 & \bfseries 2.00 \\
Q3 & 1.93 & \bfseries 2.00 & \bfseries 2.00 & 1.25 & \bfseries 2.00 \\
Q4a & 0.40 & 1.67 & 1.57 & 0.50 & \bfseries 1.83 \\
Q5a & 0.67 & 1.80 & 1.77 & 0.50 & \bfseries 2.00 \\
Q4b & 1.00 & 1.93 & 1.91 & 0.67 & \bfseries 2.00 \\
Q5b& 0.17 & 1.47 & 1.17 & 0.75 & \bfseries 2.00 \\
\hline
\bfseries FLOW & GPT 3.5& GPT 4 & GPT 4o & UG & Grad \\
\hline
Q1 & 1.00 & 1.97 & \bfseries 2.00 & \bfseries 2.00 & \bfseries 2.00 \\
Q2 & 0.43 & 1.27 & 1.30 & \bfseries 2.00 & 1.60 \\
Q3 & 0.80 & 1.57 & \bfseries 2.00 & 1.00 & 1.20 \\
Q4a & 0.17 & 0.63 & 0.57 & 1.00 & \bfseries 1.60 \\
Q5a & 0.07 & 0.13 & 0.37 & 0.00 & \bfseries 1.40 \\
Q4b & 0.76 & \bfseries 1.76 & 1.69 & 1.33 & 0.67 \\
Q5b & 0.77 & \bfseries 1.77 & 1.73 & 1.33 & 1.33 \\
\hline
\end{tabular}
\caption{Mean score out of two by population on the two tested algorithms, Euclidean (Top) and Ford Fulkerson (Bottom). The highest score(s) for each question and each algorithm are highlighted.} 
\label{tab:meanscore}
\end{figure}

\end{document}